\documentclass[11pt]{article}

\usepackage[final]{latex/acl}

\usepackage{times}
\usepackage{latexsym}

\usepackage[T1]{fontenc}

\usepackage[utf8]{inputenc}

\usepackage{microtype}

\usepackage{inconsolata}

\usepackage{graphicx}
\usepackage{xcolor}
    \usepackage{rotating}
\usepackage[table]{xcolor}
\title{Can LLMs Understand the Impact of Trauma? Costs and Benefits of LLMs Coding the Interviews of Firearm Violence Survivors}

\author{Jessica H. Zhu, Shayla Stringfield,  Vahe Zaprosyan, \\
        \bf  Michael Wagner, Michel Cukier, Joseph B. Richardson Jr. \\
        \\
        University of Maryland, College Park\\
     \small{
   \textbf{Correspondence:} \href{mailto:email@domain}{jeszhu@umd.edu}}
}

\begin{document}
\maketitle
\begin{abstract}
    Firearm violence is a pressing public health issue, yet research into survivors' lived experiences remains underfunded and difficult to scale. Qualitative research, including in-depth interviews, is a valuable tool for understanding the personal and societal consequences of community firearm violence and designing effective interventions. However, manually analyzing these narratives through thematic analysis and inductive coding is time-consuming and labor-intensive. Recent advancements in large language models (LLMs) have opened the door to automating this process, though concerns remain about whether these models can accurately and ethically capture the experiences of vulnerable populations. In this study, we assess the use of open-source LLMs to inductively code interviews with 21 Black men who have survived community firearm violence. Our results demonstrate that while some configurations of LLMs can identify important codes, overall relevance remains low and is highly sensitive to data processing. Furthermore, LLM guardrails lead to substantial narrative erasure. These findings highlight both the potential and limitations of LLM-assisted qualitative coding and underscore the ethical challenges of applying AI in research involving marginalized communities.  
\end{abstract}

\section{Introduction}

Firearm violence remains a leading cause of death among children and Black men in the United States \cite{jhopgv}. It disproportionately impacts people of color, and especially Black men. Understanding the lived experiences of firearm violence survivors through qualitative research (e.g., in-depth interviews, focus groups, participant observations) is necessary for developing and informing effective community violence interventions. Yet, since the legislation of the 1996 Dickey Amendment,  firearm violence researchers have been restricted from NIH and CDC federal funding \cite{modelfacts}. While there was a reprieve under the Biden administration, when the CDC and NIH provided \$25 million in funding for gun violence research, the Trump administration has reverted to severely restricting funding \cite{tracegv}. The long-term impact of disinvestment in firearm violence research has impacted the field’s ability to integrate new technology, like large language models (LLMs).

With the exponential growth of LLMs, qualitative coding platforms have integrated numerous AI-assisted services, from chatbots, to summarizations, to code suggestions, to complete automation, to assist along the qualitative coding process \cite{atlasti}. These services are often only accessible at an additional premium \cite{maxqda}.  Companies like OpenAI tout these LLM-powered AI automation and AI assistance tools as the starting point for the next big idea \cite{openaiyt}. However, studies have not yet demonstrated that LLMs can effectively characterize the lived experiences of minority communities, let alone those of the young Black men who are survivors of firearm violence.

In this study, we explore the value of using LLMs to code the interviews of Black men who are survivors of community firearm related violence. We bridge a gap caused by the lack of research on automated qualitative coding on diverse datasets and the limited work exploring how machine learning (ML), specifically large language models (LLMs), can support firearm violence research.  Through building a machine coding pipeline and comparing the results with human qualitative coders as ground truth, we discuss the challenges in effectively and ethically applying LLMs to research involving historically marginalized communities and firearm violence intervention.  

\section{Background}
  Qualitative methods are vital to capturing the lived experiences of individuals who have sustained violent injuries. They provide contextual insights into the social and environmental factors that contribute to violence. Qualitative coding techniques like thematic analysis enable the discovery of these insights in unstructured data. However, thematic analysis is inherently an iterative and reflective process \cite{Ahmed2025}. From re-reading texts, to identifying interesting concepts, and organizing them into themes, qualitative coding is a time-intensive, laborious process \cite{Williams2019TheAO}. This delays the publication of research findings and thereby also delays the implementation of science-backed approaches to reducing gun violence.

    Numerous studies have explored the benefits of machine learning supported qualitative coding. Beginning as early as 2008, researchers have used statistically defined natural language processing models, using word co-occurrence, latent semantic analysis, and token based topic models \cite{Dam2008, Sherin2013, Baumer2017, Crowston2012, Rodriguez2019, Lennon2021} to explore qualitative data, sometimes incorporating interactive human collaboration \cite{collabcoder, Gebreegziabher2023}.  These approaches all demonstrated the potential for speedups in the qualitative coding process through the assistance of natural language processing and machine learning, especially for short texts \cite{Feuston2021}. However, these initial approaches still required various degrees of data labeling, rule development, parameter tuning, retraining, and additional human interpretation. The lack of transparency and potential for an overwhelming number of code recommendations was also found to detract from machine learning integration in qualitative coding \cite{Rietz2021, Lennon2021, Marathe2018}.
    
    With the advent of AI agents and generative LLMs, researchers and platforms are experimenting with using LLMs to fully automate the qualitative coding process with minimal human interaction. LLMs have been found to be helpful for deductive coding techniques \cite{ziems2024, Ranjit2025}. LLMs have also found success in thematic analysis and inductive coding in scenarios like consumer product surveys \cite{Dai2023}, online forums \cite{NagarajRao2025, Sharma2025}, quotes from social science studies \cite{Parfenova2025}, litigation documents \cite{Zhong2025}, and mental health interviews of healthcare professionals \cite{Singh2024}. All of the aforementioned studies except \cite{Parfenova2025, Zhong2025}, used closed-source models with over 100B parameters. These services and model size are both financially and computationally inaccessible for low-resourced, historically marginalized communities. In addition, these studies were primarily short form texts from online forums, surveys, or discussions in standard English, which is not representative of the long form interviews needed to understand and address the disproportionate impact of firearm violence on young Black men.
    
    This under-served population has been substantially biased against by ML and LLMs. There have been few ML applications for firearm violence intervention published despite the boom in ML research applied to other public health fields \cite{modelfacts}. While LLMs have the potential to be beneficial to qualitative coding in social work research, they remain significantly limited by their potential for hallucinations and bias against BIPOC \cite{Patton2023}. LLMs have been found to generate covertly racist decisions against African American English (AAE). Reinforcement learning with human feedback often exacerbates implicit biases while decreasing explicit biases \cite{Hofmann2024}. Even guardrails, which are ostensibly supposed to protect users, have demonstrated biases against certain identities \cite{Li2024}. While there is debate as to the actual harms of implicit biases, research has found implicit biases to manifest as systemic racism \cite{Galvan2024}. To mitigate the propagation of systemic racism, researchers and developers must conduct a thorough analysis of LLMs on diverse datasets in partnership with communities before allowing LLMs to be a ubiquitous part of thematic analysis.

\section{Methodology}
\begin{figure}[ht]
        \centering
        \includegraphics[width=1\linewidth]{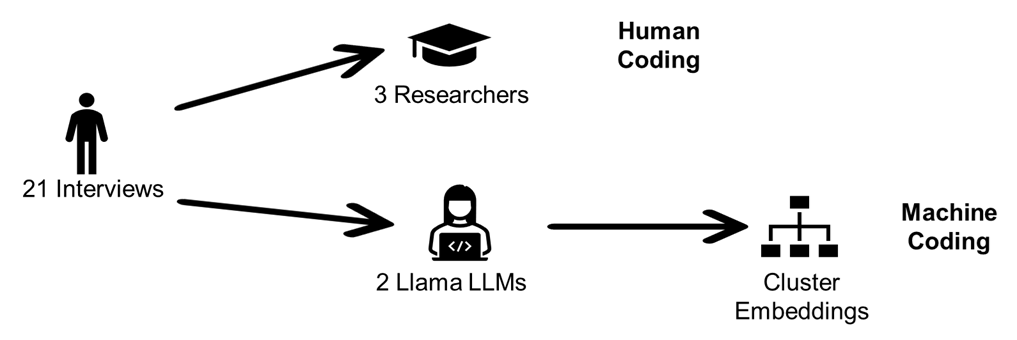}
        \caption{Our pipeline for analyzing interviews and creating codes}
        \label{fig:pipeline}
    \end{figure}

    In an effort to resolve one facet of concerns about LLM biases and effectiveness, we partnered with community violence intervention researchers to interrogate the effectiveness of LLMs in understanding the lived experiences of Black firearm violence survivors. We compare the effectiveness of language models in identifying codes in the interviews of 21 firearm violence survivors with the codes identified by qualitative researchers (see Figure \ref{fig:pipeline}).

    \subsection{Data}
     We analyze the interviews of 21 Black/African-American men that were previously collected between January 2013 and December 2015 with approval from the IRB (protocol number 343085-1). These subjects were identified after being admitted to a Level 2 trauma center in the DC metropolitan area for a violent injury. They were asked to be volunteers following their admission and were compensated with \$50 per interview. Dr. Richardson, an experienced African American studies anthropologist and firearm violence researcher, led one on one interviews following a semi-structured protocol. Interviews lasted approximately 60 minutes and were conducted in a private room.  The interviews were later manually transcribed and de-identified by two research assistants. Transcribers were undergraduate students and identified as white men and women. Once transcripts were cleaned, participant names were changed to pseudonyms prior to data analysis. The principal investigator kept a master spreadsheet with participant names and pseudonyms in a password protected folder.

    \subsection{Human Coding (HC)}
    Interviews were coded manually using qualitative software. The AI coding functionality in the software was not used by the team. Human coders  followed grounded theory and used inductive thematic analysis coding techniques \cite{Ahmed2025} to identify initial codes. They identified as Black/African-American women, with one undergraduate student and two research staff. One coder had more than two years of experience. 
    
    Human coders were instructed to review the study protocol, which focused on specific experiences around high-risk sexual activity, previous trauma history, perceived community violence, and feelings of retaliation, prior to coding. They were encouraged to highlight codes related to the initial focus, as well as any codes that were not a part of this initial objective. Once the human coders identified the initial codes, member check-in strategies were used to review the accuracy of the codes. Initial human annotated codes (HC codes) were then refined and merged into formal HC codes. Formal codes were reviewed by the principal investigator for accuracy prior to finalization.

    \subsection{Machine Coding Pipeline (MC)}
      We also built a machine coding pipeline using open-source models to evaluate the effectiveness of more accessible low-resourced LLMs (those that can be run on relatively affordable hardware) at automated qualitative coding. We first cleaned and processed the long form interviews into machine readable chunks before passing them to the LLMs. We experimented with open-source models running on university research servers to protect the sensitive interview data. Experiments were conducted on one 40GB A100 node, for 8B parameter LLMs; one 24GB L4 GPU, for 1B parameter LLMs; and a 16GB RAM laptop for all Sentence Transformer language models. Code is available at \url{https://github.com/jhzsquared/AIvsHumanCoding}.

    \subsubsection{Data Processing}
     We removed auxiliary information from the transcripts (e.g., time stamps, additional commentary on background noise) and extracted text by speaker (interviewer or subject). Due to computational resource limitations when using larger models, we chunked the data into a maximum of 256 token length sections. We used 256 tokens to approximate the length of a paragraph’s worth of discussion on a given topic. We experimented with ``paired chunks’’ where we sequentially separated interviews such that every line from an interviewer was first paired with the following subject’s turn and then split further only if it was too long. 
    For ``question chunks’’, we extracted the questions from the original interview protocol. We then encoded the questions as well as every subject response turn. The embeddings from two Sentence Transformer models (``all-mpnet-base-v2'' and ``multi-qa-MiniLM-L6-cos-v1'' \cite{reimers-2020-multilingual-sentence-bert}) were concatenated to form a final ensembled embedding. We then calculated the cosine similarity between the embedded questions and responses. Responses were assigned to the question (or an ``other’’ category) where they had a maximum similarity score, as long as they met a 20\% similarity threshold. This threshold and the models were selected from observational checks of the 30 highest scoring matches from the question-response pairs after testing different thresholds and models. Responses were further split into at most 256 length token chunks. 
    
    For the smallest LLM (1B parameters), we also experimented with passing each interview in their entirety (``full text''). We did not have sufficient computational capacity available to do the same for the larger, 8B parameter LLM. A breakdown of the data's descriptive statistics is shown in Table \ref{tab:datastats}.
    
    \begin{table}[h]
        \centering
        \begin{tabular}{|l l|}
           \hline
        
        \# of words & 11503 (SD=4955) \\
        per interview & \\
        \hline
        \# of words  & 34 (SD=64) \\
        per response & \\
           \hline
        \# of response turns  & 481 (SD=173) \\
           per interview & \\
        \hline
        \# of paired chunks & 249 (SD=92) \\
           per interview & \\
           \hline
        \# of question chunks & 30 (SD=35) \\
        per question & \\
           \hline
        Total \# of questions  & 28 \\
        in the protocol & \\
           \hline
        Total \# of interviews & 21 \\
           \hline
    \end{tabular}
      \caption{Average and standard deviation of various descriptive statistics of the corpus of semi-structured interviews}
     \label{tab:datastats}
    \end{table}
    
    \subsubsection{Code Generation}
    For initial code generation, we used Llama-3.2-1B-Instruct (Llama 1B) and Llama-3.1-8B-Instruct (Llama 8B) \cite{llama3}. We selected these models for their consistently high performance on standard benchmarks for instruction based tasks within the confines of our compute limitations and to evaluate the impact of larger model sizes and more flexible resources. To extract relevant themes from the preprocessed interview text, we passed zero-shot prompts while varying the identity the system should assume and the context of the queries (see Appendix \ref{sec:appendixprompts} for additional prompt settings). We used the term ``themes’’ rather than ``codes’’, as during initial tests, we found that the models were more helpful with the term ``themes’’. The word ``codes’’ often resulted in output discussing coded language (i.e., slang) rather than qualitative codes. Figure \ref{fig:exampleprompt} shows an example of one of the many prompts tested.

    \begin{figure}[ht]
        \textbf{System}: ``You are \colorbox{pink}{an African American Studies} 
        \colorbox{pink}{anthropologist} analyzing interviews to understand \colorbox{yellow}{the experiences of gun violence survivors}. Your response should be a numbered list with each item on a new line.'' \\    
        \textbf{User}: ``List the themes observed in the following interview excerpt:"
         \\
         \{INTERVIEW EXCERPT\}''
        \caption{Example prompt -- The first highlighted portion is the ``identity'' we have the system assume, and the second highlighted portion is the ``context'' that is provided.}
        \label{fig:exampleprompt}
    \end{figure}
    
       After the initial machine generated (MC) codes were generated, in mirroring the process of thematic analysis, we clustered the output codes into ``formal codes’’. We parsed the initial LLM output into individual codes, aggregated all results,  and de-duplicated them. We then used BERTopic \cite{grootendorst2022bertopic} to cluster the initial codes into interpretable ``formal'' codes. We embedded the initial codes with  ``all-mpnet-base-v2’’, the top performing pretrained Sentence Transformer model \cite{reimers-2020-multilingual-sentence-bert}. We used the topic model that maximized silhouette score after conducting grid-search over key hyperparameters. Parameter details can be found in Appendix \ref{sec:params}. We then prompted the originating LLM to generate formal MC code names using each topic model cluster’s representative keywords and its three most representative originating codes along with the LLM generated justification for those codes. See Appendix \ref{sec:appendixprompts} for the prompt used to generate the formal codes.

    \subsubsection{Evaluation}
   To evaluate the quality of the machine codes, we introduce two metrics: Percent Captured and Percent Relevant. Percent Captured is the percentage of the formal human annotated codes with a machine code match. It reflects how well the machine automated coding pipeline extracts what the human annotators deemed to be important and could be considered a fuzzy recall score. Percent Relevant is the percentage of MC codes that match any HC codes (initial or formal). If there are many MC codes generated, the odds of a code matching with an HC code is higher, but the overall relevance of the output will be lower. While it could be considered analogous to a fuzzy precision score, Percent Relevant does not necessarily indicate that the additional MC codes are hallucinations or not useful. These ``irrelevant'' MC codes could also be novel codes that the human annotators did not think of or did not deem important.
   
    To calculate these scores, we estimated the semantic similarity between the machine codes and human codes using the cosine similarity of their embeddings from ``all-mpnet-base-v2’’ \cite{reimers-2020-multilingual-sentence-bert}. While human validation would be ideal, due to the intractable number of codes the LLM often generated and the numerous experiments completed, this was more time efficient. An MC code was considered to be a match with an HC code if its cosine similarity score was greater than 0.6. This threshold was determined through reviewing a sample of initial matches at varied thresholds. We evaluated both the initial and formal machine generated codes using these metrics.


\section{Results}
   The human coders identified 41 initial codes and 11 formal codes across all 21 interviews. They  spent an average of 12 hours each coding. Depending on the parameters, initial MC generation took 3.4 hours (SD = 3.3) on average over the 118 experiments that we ran. The large variation in time was primarily a factor of the data processing technique: the full interview experiments took 10 minutes on average, paired chunks took 7.3 hours on average, and question chunks took 2.5 hours on average. The formal code's topic model pipeline took a trivial amount of time (less than 5 minutes on average). Statistics and scores from the best models are further summarized in Table \ref{tab:resultstats}. Their parameter settings are described in Appendix \ref{sec:params}

    \subsection{Initial MC Output}
\begin{figure}[ht]
        \centering
        \includegraphics[width=1\linewidth]{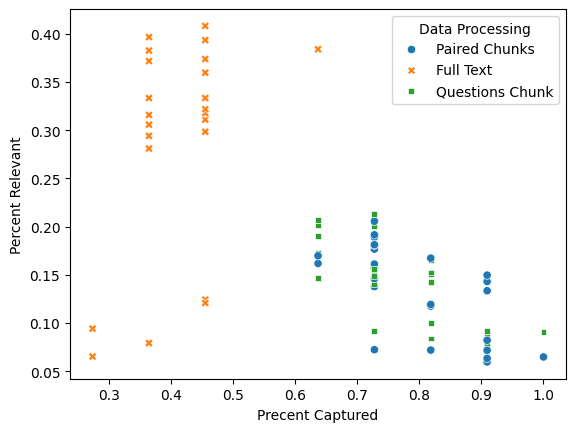}
        \caption{Percent Captured versus Percent Relevant of initial MC codes by data processing techniques using Llama 1B}
        \label{fig:dataprocscorel1b}
    \end{figure}

   Prior to clustering, the MC pipeline generated orders of magnitude greater initial codes than the human coders. On average, 3072 (SD=2838) unique codes were identified from the initial LLM prompts. As few as 98 codes were generated using the full text data processing strategies and as many as 16000 for the paired chunk strategies. However, the full text strategy's initial Percent Captured was never as high as the other data processing strategies though it had significantly higher relevance (see Figure \ref{fig:dataprocscorel1b}). Full text output's Percent Relevance was 11\% and 12\% higher on average than both question chunks and paired chunks, respectively. Paired chunks had significantly higher Percent Captured than questions chunk and full text, outperforming them by 7\% and 32\% on average, respectively, per a Wilcoxon signed-rank test at .05 p-value. Question chunks had a statistically significant higher Percent Relevance than paired chunks, but only by a difference of 1\%.
   
   Specifying the identity to be ``an African American studies anthropologist'' rather than ``a Black anthropologist'', had a marginal, though statistically significant difference, of 0.4\% in Percent Relevance. Specifying the context to be ``the experiences of Black men as gun violence survivors'' rather than just ``the experiences of gun violence survivors'' also resulted in a statistically significant 1\% higher average Percent Captured, though there was no discernible impact on Percent Relevance.

      In comparing model size, Llama 8B outperformed Llama 1B by an average of 2.6\% based on Percent Relevance, which was statistically significant per a Wilcoxon test at .05 p-value. Llama 8B did not outperform Llama 1B by Percent Captured, with Llama 1B actually outperforming (not statistically significantly) on average by 1.8\%. We found that the best initial MC codes from either model were equally well aligned with the formal human codes, with 100\% of the formal HC codes captured using both Llama 1B and Llama 8B (see Appendix \ref{sec:appendixout}, Table \ref{tab:initialcodes} for the alignment of best and worst MC output to formal HC codes). However, the worst Llama 1B output had much lower scores than the worst Llama 8B experiment's (Table \ref{tab:resultstats}).
      
      Overall, Percent Captured was as low as 27\% (average of 71\% and SD 18\% across all experiments). Percent relevance was consistently low for all settings, with an average of 10\% (SD = 5\%) across all experiments. The best and worst outputs all identified codes related to ``Masculinity'' and ``Perceived community violence'', while only the best outputs also coded ``Altercation leading to hospital visit'',   ``Behavior changes after incident'',  ``Reasoning for beef starting'', and ``Sexual activity''.

    \subsection{Formal MC Output}
   After evaluating the Percent Captured and Percent Relevant of the initial MC output, we selected the best and worst results from each Llama model's experiments for formal code generation. We passed these four sets of machine generated codes through the clustering pipeline. The specific prompts and data processing settings for these experiments are in Appendix \ref{sec:appendixout}. Clustering the output greatly reduced the number of codes down to numbers close to the initial HC results (Table \ref{tab:resultstats}).

    \begin{table*}[ht]
        \centering
        \begin{tabular}{|l | l| l| l| l| l|}
        \hline
         & \textbf{HC} & \textbf{Best L1B} & \textbf{Worst L1B} & \textbf{Best L8B} & \textbf{Worst L8B} \\
        \hline
        \textbf{Time Spent (hrs)} & 35 & 1.45 & .16 & 5.75 & .63 \\
        \hline
        \textbf{\# of Initial Codes} & 41 & 2997 & 184 & 6968 & 690 \\
        \hline
        \textbf{\# of Formal Codes} & 11 & 57 & 15 & 45 & 54 \\
        \hline
        \textbf{Silhouette Score} & N/A & .694 & .785 & .68 & .73 \\
        \hline
        \textbf{\% Captured} &  & & & & \\
        \quad Initial & N/A & 100\% & 27\% & 100\% & 64\% \\
        \quad Formal & N/A & 36\% & 0\% & 36\% & 45\% \\
        \hline
        \textbf{\% Relevant} & & & & &  \\
        \quad Initial & N/A & 9.0\%  & 6.5\% & 13.8\%  & 19.7\% \\
         \quad Formal & N/A & 28\% & 0\% & 28\%  & 36\%\\
        \hline
    \end{tabular}
    \caption{Statistics from the best and worst Llama 1B and Llama 8B experiments. Model settings are described in Appendix \ref{sec:appendixout}.} 
    \label{tab:resultstats}
    \end{table*}
    
   While one may expect the quality of the codes to improve with aggregation, we found that clustering resulted in formal MC codes that were less aligned with the HC codes than the initial MC codes were. This is seen in the notable drop in Percent Captured between initial and formal codes in Table \ref{tab:resultstats}. This drop occurred regardless of the originating output's scores. Clustering rebalanced the difference between the best Llama 8B and the worst Llama 8B's output. It also neutralized the difference between the Best Llama 1B and the Best Llama 8B. After transforming the formal HC codes with each result's pretrained topic model, we found that many of them were aligned with clusters where they lacked semantic similarity to the representative cluster name (i.e., the formal MC code). This indicates that in spite of their high silhouette scores, the clusters still encompass a wide breadth of codes. Only for a few human codes do their aligned clusters also have high semantic similarity. For example, for the formal HC code ``Masculinity'', the best Llama 1B result's topic model aligned it to the formal MC code ``Dealing with masculinity and identity'', which was a semantic similarity match (per cosine similarity using ``all-mpnet-base-v2''). However, the best Llama 8B result's topic model aligned ``Masculinity'' to the formal MC code ``Racial profiling and police bias'', even though it was most semantically similar to ``Role models in masculinity''. Appendix \ref{sec:appendixout}, Table \ref{tab:formalcodes} displays the alignment of all formal HC codes to clusters versus semantic match. Meanwhile, Llama 1B's worst output fixated on non-standard English linguistic characteristics, like the usage of the words ``get'' and ``like''. None of its formal MC codes matched with HC codes based on semantic similarity.
   
\begin{figure*}[ht]
        \centering
        \includegraphics[width=.75\paperwidth]{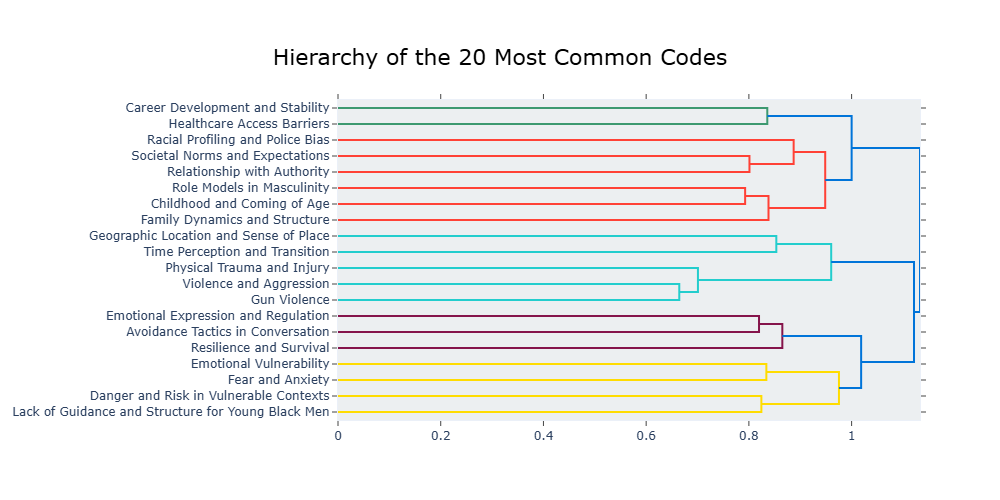}
        \caption{The relationship between the top 20 most common codes from the best performing Llama 8B experiment}
        \label{fig:hierplottop}
    \end{figure*}

  Clustering did substantially improve the overall relevance for all outputs (disregarding the worst Llama 1B since it had no matches after clustering). It reduced hundreds and thousands of initial codes to fewer than 60. Clustering also highlights potential new insights and relationships between codes. From the hierarchical structure that resulted from clustering the best Llama 8B's codes (Figure \ref{fig:hierplottop}), we can observe topics structured around employment and healthcare; social norms and relationships; descriptions of the source of injury; and varying facets of emotions. These interrelationships provide new insights into these narratives.
    
  We also examined the clusters that did not align with any formal HC codes to explore if the low Percent Relevance is a result of novel discoveries or hallucinations. We highlight some of the new codes from the best performing experiments in Table \ref{tab:weirdcodes}. With our MC pipeline, we are able to recall the interview excerpts that have led to each code. Upon reexamination of a sample of the most representative codes' source data, we found that many of these new MC codes often hold a tenuous relationship with the source text, such that the LLM appears to have read between the lines. The new codes from Llama 8B were generally more justified by the representative source text than those from Llama 1B. For example, the code ``Internet'' from Llama 1B is partially justified, as they were discussing ``World Star Hiphop''. While ``Gang affiliation codes'' was not justified at all since the representative source texts are primarily discussing sources of injury and fights. On the other hand, ``Racial profiling and police bias'' from Llama 8B is likely valid based on the discussions on unfair trespassing charges and differences in police treatment in neighborhoods. Meanwhile, ``Public perception of the criminal justice system'' is not clearly justified. Even though the interviews did discuss the subject's experiences with the criminal justice system and how they perceive police, the representative texts do not clearly justify coding it as ``public perception''.  More experimentation and validation with humans in the loop is needed to determine if this discrepancy between Llama 1B and Llama 8B is a consistent trend. 
     
    \begin{table*}[ht]
    \centering
    \begin{tabular}{| l  l l l |}
        \hline
        \textbf{Llama 1B} & \textbf{Valid} & \textbf{Llama 8B} & \textbf{Valid}  \\
        \hline
         Continuation & N & Stigma and Shame in Vulnerable Populations & M\\
        \hline
        Checkup & N & Social Commentary and Moral Judgment & M \\
        \hline
        Perpetrator-Victim Dynamics & N & Mental Health and Wellbeing of Black Men & Y \\
        \hline
        Frustration and Anger & N & Racial Profiling and Police Bias & Y \\
        \hline
        Perpetrators & N & Geographic Location and Sense of Place & M \\
        \hline
        Gang Affiliation Codes: Blood Gangs & N & Public Perception of the Criminal Justice System & M\\
        \hline
        Internet & M & Self-Awareness and Personal Growth & Y \\
        \hline
    \end{tabular}  
    \caption{New machine generated formal codes from the best performing experiments (Y= sufficient data to support it; M=data may support it; N=insufficient data to support it)}
    \label{tab:weirdcodes}
    \end{table*}
        
    \subsection{LLM Refusals}
\begin{figure} [ht]
        \centering
        \includegraphics[width=1\linewidth]{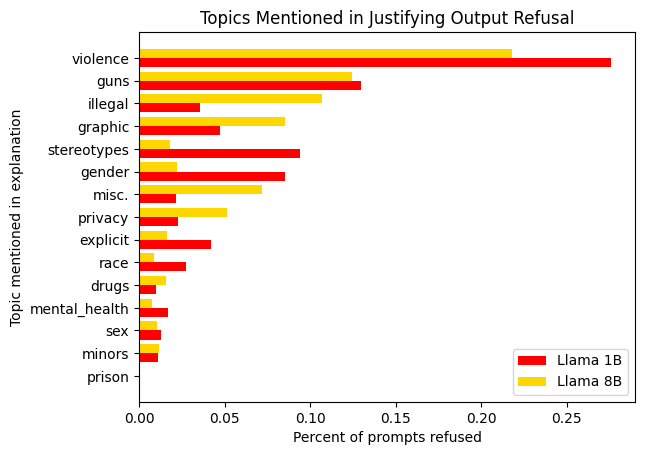}
        \caption{Distribution of justifications for LLM refusals. Note: Justifications may be aligned with multiple topic areas (e.g., firearm violence would count for both gun and violence).}
        \label{fig:refusals}
    \end{figure}

      Although numerous codes were generated, we found that on average, 44\% (SD = 25\%) of each prompt request refused to generate output. Refusals also occurred during clustering naming. The primary reasons the LLMs used to justify refusal are displayed in Figure \ref{fig:refusals}. These were determined through a regular expression of keywords related to each category. These keywords are listed in Appendix \ref{sec:params}. The primary justification LLMs provided for refusal was the descriptions of the survivor's firearm violence experiences, which were often deemed too graphic. In some cases, the LLMs refused to ``discuss content that promotes or glorifies violence''. There were cases where they also refused ``to sexualize or objectify survivors of firearm violence''. The LLMs deemed the descriptions and questions on the subject's experiences were too graphic, even though we explicitly stated that we are requesting themes in our capacity as researchers. In addition, explicit and AAE language in the interviews, especially the use of the n* word, almost guaranteed refusal. This is despite the text not being derogatory. Discussions regarding sexual activity and race were also often refused, which is problematic as these topics were focal points of the original study.  These interviews research not only the factors leading to traumatic events, but also if young Black men at high risk for violent victimization may also be engaged in high risk sexual behavior. Due to LLM refusals, much of the narrative discussing these research questions is ignored.
      
   It is important to note that we did not find a correlation between the percent refused per experiment and the number of outputs generated, nor with our quality metrics: Percent Relevant and Percent Captured (Pearson's correlation coefficient of $< |.2|$). This demonstrates that the formal MC codes we successfully captured were substantially present in the data that was not ignored. It also leads to the question of what additional, informative codes exist within the 44\% of data that was erased. 

\section{Discussion}
    These results show that while a low-resourced fully automated pipeline has the potential to extract codes, the results are inconsistent. The best MC experiments effectively extracted the formal human codes with minimal tuning and additional background. However, there were significant differences in the quality of the initial MC codes with minimal changes to the data processing techniques, core prompt, and identity.

   The MC pipeline also initially generated an intractable number (on average 3000) of unique codes for human analysis, defeating the time saving goals of automation. But with the inclusion of a second clustering step, the machine pipeline generated a reasonable (fewer than 100) number of human readable codes. Yet these ``formal’’ MC codes appeared to have lower validity overall than the initially generated codes, as well as a greater likelihood of highlighting hallucinated, misrepresented, and/or stereotypical codes, as shown by our Percent Captured and Percent Relevant scores.

    These LLMs also were not difference-aware \cite{Wang2025}. Their guardrails were biased \cite{Li2024}. Our results indicate that not only are LLMs biased against AAE, but they are also biased against the experiences of firearm violence survivors, especially the traumas they have survived. These biases and misinformed guardrails resulted in narrative erasure of the interviews, where up to 65\% of some subjects’ experiences were ignored. While Llama is known for being more safety minded against controversial prompts \cite{UGIleaderboard}, these findings demonstrate significant concerns about the efficacy of LLMs that are marketed to assist all qualitative research, to include marginalized and historically under-served populations, like those impacted by firearm violence. 

      Though human coders come with their own biases, unlike our automated pipeline, they were able to understand AAE and not fixate on colloquialisms. While there is an emotional burden to reading traumatic experiences, they successfully extracted codes relevant to the study protocol and were not stopped by the  depictions of violent injuries pertinent to this study. However, they needed over 10 times as many hours to code, which demonstrates the significant potential for time savings if models are accessible, reliable, and ethically aligned.
      
 Given the resource and especially computational constraints that many minority communities and positive impact organizations face, it is promising that Llama 1B, which required on average 6GB of VRAM, nears Llama 8B's performance, which required 34GB of VRAM to run our pipeline. However, Llama 1B's performance was less reliable and more impacted by fixations on colloquialisms characteristic of AAE. Regardless of the model, automated pipelines will still require human experts to validate codes. Luckily, as demonstrated by our novel code investigation, the validation time can be reduced through effective data tracking strategies combined with more interpretable clustering techniques like BERTopic. Nevertheless, significantly more adjustments will still be needed for reliable, unbiased inductive coding.
  
\section{Conclusion}
    Our research demonstrates that an automated pipeline for coding long-form interviews still has significant limitations. Through implementing open-source pre-trained instruction tuned models, our automated pipeline demonstrated high Percent Captured using as few as 12 GPU hours. However, more low-resourced methods are needed to support qualitative coding of non-short form, survey based, standard English data. We cannot rely on AI qualitative coding tools to understand the lived experiences of diverse and vulnerable communities, without significant improvements to LLMs. Though an automated coding pipeline may appear faster, it cannot yet match the consistent quality of human coding. The time spent evaluating machine generated codes rapidly devalues any initial time savings. If one further factors in the time to tune and train an AI pipeline, the time gains are drastically reduced.

    We currently cannot use AI assistants and LLMs to code these narratives. There remains a significant risk of narrative erasure as well as non-representative and hallucinated results. Methods are needed that can efficiently increase coding quality and reliability, and decrease the risks of misrepresentation and implicit biases for low-resourced settings. As our research shows, larger models may not even offer a substantial improvement in performance, despite their significant increase in computational, financial, and environmental burden on an already underfunded research community. Therefore, we must be deliberate in building low-resourced AI tools that represent minority communities, like the young Black men disproportionately impacted by firearm injuries. 

\section*{Limitations}
    This study draws from subjects in the Washington, DC area. As such, they represent only one of many AAE regional patterns. In addition, interviews were manually transcribed. While transcriptions were further reviewed by researchers fluent in AAE, there is still a potential for human error. Our prompting strategy was also limited and did not include the study protocol as we wanted to minimize the potential for the LLM to generate themes from the protocol rather than the interview. Including the study protocol would otherwise lead to a more informed context, assuming the LLM has sufficient capacity for longer inputs. We minimized the potential discrepancies in human versus machine codes, by asking the human coders to code items not relevant to the initial study objective as well. Further human validation of our evaluation and data processing pipeline would be beneficial prior to generalizations. However, through our limited samples, we believe our thresholds to be sufficiently reliable for this study. We also recognize that further examination of LLMs in other model families, or of larger sizes with quantization would be beneficial. We used full precision Llama 3.2 1B and 3.1 8B due to resource limitations, the widespread usage of Llama, and to avoid additional variations from quantization. 

\section*{Ethical Considerations}
    Aside from the potential costs of LLM automated coding already discussed (e.g., narrative erasure, misrepresentation, and stereotyping), there is a risk in using these LLM generated output code to generalize the experiences of gun violence survivors. While we do share the final formal codes from MC, further validation of the MC pipeline is needed for these codes to be a reliable source of qualitative analysis. Data will not be released to protect interview subjects. To further ensure privacy, we have manually validated that none of the output codes are identifiable. 
    
\section*{Acknowledgments}
    Thank you to Hannah Balconoff and Bailey Skeeter for their work coding the interviews and the undergraduate researchers who supported transcriptions. The authors would also like to acknowledge the University of Maryland supercomputing resources (https://hpcc.umd.edu) made available for conducting the research reported in this paper as well as the support of various University of Maryland College Park grants, to include their Summer Research Fellowship, School of Public Health's Prevention Research Center Seed Grant Program, Consortium for Race, Gender, and Ethnicity: Faculty Seed Grants for Developing Qualitative Work, College of Behavioral and Social Sciences Dean's Research Initiative, and Research and Scholarship Award.

\bibliography{latex/custom}

\appendix
\section{Prompts}
\label{sec:appendixprompts}
The full list of prompts, identities, and context experimented with are as follows.

\subsection{Identities}
\begin{itemize}  
    \item an anthropologist
    \item an African American Studies anthropologist
    \item a Black anthropologist
   
\end{itemize}

\subsection{Contexts}
\begin{itemize}
    \item the experiences of gun violence survivors
    \item the experiences of Black men as gun violence survivors
\end{itemize}

\subsection{Prompts}
Prompts by data processing technique:

\textbf{For paired chunks:}
    \begin{itemize}
        \item \textbf{base theme:}  ``user'': ``What themes are observed in the following interview excerpt? Your response should be a numbered list with each item on a new line. \\ 
        \{INTERVIEW\}''
        \item \textbf{base\_t:} ``system'': ``You are \{IDENTITY\} analyzing interviews to understand \{CONTEXT\}. Your response should be a numbered list with each item on a new line.'' \\
        ``user'': ``List the themes observed in the following interview excerpt: \\
        \{INTERVIEW\}''
    
        \item \textbf{cot\_t:} ``system'': ``You are \{IDENTITY\} analyzing interviews to understand \{CONTEXT\}. Your response should be a numbered list with each item on a new line.'' \\
       ``user'': "List the themes observed in the following interview excerpt. Provide quotes from the interview that demonstrate the themes. \\
       \{INTERVIEW\} ''
    
        \item \textbf{``base\_c'':} ``system'': ``You are \{IDENTITY\} applying inductive coding techniques to understand \{CONTEXT\} from interview data. Your response should be a numbered list with each item on a new line.'' \\
       ``user'': ``List the codes observed in the following interview excerpt: \\
       \{INTERVIEW\}''

        \item \textbf{``novel\_cot\_t'':} ``system'': ``You are \{IDENTITY\} analyzing interviews to understand \{CONTEXT\}. Your response should be a numbered list with each item on a new line.'' \\
        ``user'': ``List the novel themes observed in the following interview excerpt. Provide quotes from the interview that demonstrate the themes. \\
        \{INTERVIEW\}"
    
    \end{itemize}
 \textbf{For question chunks:}
        \begin{itemize}
            \item \textbf{base\_theme}   
             ``user'': ``What themes are observed in the following interview responses to the question: \{QUESTION\}? Your response should be a numbered list with each item on a new line. \\ \{INTERVIEW\}''            
            \item \textbf{base\_t:}           
            ``system'': ``You are \{IDENTITY\} analyzing interviews to understand \{CONTEXT\}. Your response should be a numbered list with each item on a new line.'' \\
           ``user'': ``List the key themes observed in the following interview responses to the question: \{QUESTION\}:\textbackslash{}n \{INTERVIEW\}''            
            \item \textbf{cot\_t:}            
            ``system'': ``You are \{IDENTITY\} analyzing interviews to understand \{CONTEXT\}. Your response should be a numbered list with each item on a new line.'' \\
           ``user'': ``List the key themes observed in the following interview responses to the question: \{QUESTION\}. Provide quotes from the interview that demonstrate the themes. \\ \{INTERVIEW\}''
                       
            \item \textbf{base\_c:}            
            ``system'': ``You are \{IDENTITY\} applying inductive coding techniques to understand \{CONTEXT\} from interview data. Your response should be a numbered list with each item on a new line.'' \\
           ``user'': ``List the codes observed in the following interview responses to the question: \{QUESTION\}. Responses: \\ \{INTERVIEW\}''            

           \item \textbf{novel\_cot\_t:}    
            ``system'': ``You are \{IDENTITY\} analyzing interviews to understand \{CONTEXT\}. Your response should be a numbered list with each item on a new line.'' \\
            ``user'': ``List the novel themes observed in the following interview responses. Provide quotes from the interview that demonstrate the themes. \\ \{INTERVIEW\}''
         \end{itemize}
 \textbf{For full text:}
        \begin{itemize}
            \item \textbf{base\_theme:}  
            ``user'': ``What themes are observed in the following interview? Your response should be a numbered list with each item on a new line.\\
            \{INTERVIEW\}''
        
            \item \textbf{base\_t:} 
            ``system'': ``You are \{IDENTITY\} analyzing interviews to understand \{CONTEXT\}. Your response should be a numbered list with each item on a new line.'' \\
       ``user'': ``List the themes observed in the following interview: \\
       \{INTERVIEW\}''

            \item \textbf{cot\_t:} 
            ``system'': ``You are \{IDENTITY\} analyzing interviews to understand \{CONTEXT\}. Your response should be a numbered list with each item on a new line.'' \\
            ``user'': ``List the themes observed in the following interview. Provide quotes from the interview that demonstrate the themes. \\
            \{INTERVIEW\}''
        
            \item \textbf{base\_c:}
        ``system'': ``You are \{IDENTITY\} applying inductive coding techniques to understand \{CONTEXT\} from interview data. Your response should be a numbered list with each item on a new line.'' \\
       ``user'': ``List the codes observed in the following interview:\\
       \{INTERVIEW\}"       
            \item \textbf{novel\_cot\_t:}        
        ``system'': ``You are \{IDENTITY\} analyzing interviews to understand \{CONTEXT\}. Your response should be a numbered list with each item on a new line.'' \\
       ``user'': ``List the novel themes observed in the following interview. Provide quotes from the interview that demonstrate the themes. \\
       \{INTERVIEW\}''
        \end{itemize}

\textbf{To get the topic name}: 

``system'': ``You are an assistant that extracts high-level topics from texts. Only return the topic name.''

``user'': This is a list of texts where each collection of texts describe a topic. After each collection of texts, the name of the topic they represent is mentioned as a short-highly-descriptive title. \\
--- \\
Topic: \\
Sample texts from this topic: \\
    - {DOCUMENTS} \\
Keywords: {KEYWORDS} \\
Topic name:''

The DOCUMENTS are the originating code and justification (as relevant). The KEYWORDS are outputs of BERTopic.

\section{Pipeline Parameters}
\label{sec:params}
 We used a temperature of 0.6 and top\_p of 0.9 for both Llama 3.1 8B (L8B) and 3.2 1B  (L1B) Instruct models to support more creative but still reliable output. 
 The prompt settings and topic modeling parameters we found to have the best and worst codes are shown in Table \ref{tab:modelparams}

\begin{table*}[ht]
\centering
\begin{tabular}{| l |p{2.5cm}| p{2.5cm}|p{2.5cm}|p{2.5cm}|}
\hline
 & Best L1B & Worst L1B & Best L8B & Worst L8B \\
\hline
Data processing & base\_c & base\_c & base\_c & cot\_t \\
\hline
Prompt & Question chunks & Full text & Paired chunks & Question chunks \\
\hline
Identity & African American Studies anthropologist & Black anthropologist & Black anthropologist & African American studies anthropologist \\
\hline
Context & the experiences of gun violence survivors & the experiences of gun violence survivors & the experiences of Black men as gun violence survivors & the experiences of Black men as gun violence survivors \\
\hline
n\_neighbors & 15 & 5 & 20 & 5 \\
\hline
n\_components & 5 & 5 & 5 & 2 \\
\hline
min cluster size & 15 & 5 & 40 & 5 \\
\hline
\end{tabular}
\caption{Pipeline parameters resulting in best and worst LLM codes}
\label{tab:modelparams}
\end{table*}

To analyze the source of refusals, we relied on regular expressions. We did experiment with topic modeling the refusal sentences as well, but the sentences were too noisy for well structured and interpretable clusters.

The terms used in the regular expressions for each  refusal category are as follows:
\begin{itemize}
    \item illegal: illegal, criminal, crime
    \item violence: violent, violence, war, brutality
    \item guns: firearm, shot, gun, shooting
    \item explicit: explicit, n-word, profane, profanity, obscenity, nigga
    \item stereotypes: hate, speech, derogatory, stereotype, slur, discriminatory, discriminate, stigma
    \item mental\_health: mental, suicide, crisis, self-destructive
    \item graphic: dangerous, graphic, disturbing, harmful
    \item sex: sex, condom, HIV, AIDS, std
    \item drugs: drug, marijuana, substance abuse, weed
    \item gender: women, gender, man, men, woman, female, male
    \item race: black, African, racial
    \item minors: child abuse, child, minor, children
    \item privacy: identify, personal, individual, medical
    \item prison: jail, prison, justice system, incarceration
    \item misc: all sentences that did not fall into one of these categories
\end{itemize}

\section{Initial HC and formal MC output}
\label{sec:appendixout}
    We share the initial human annotated codes as well as the formal output codes from the best and worst Llama 1B and Llama 8B pipeline results to allow further comparison of the output. We have manually removed duplicates, topic name refusals, and ``null'' topic. Duplicated topic names are marked with ``(number of duplicates)''. Given the systemic errors in pipeline results, the MC outputs should not be directly used to draw conclusions on the lived experiences of firearm violence survivors.
    
    \begin{table*}[ht]
    \centering
        \begin{tabular}{| p{3cm} | p{3cm}| p{3cm} | p{2cm} |p{3cm} |}
        \hline
        HC & Best L1B & Best L8B & Worst L1B &  Worst L8B \\
        \hline
        Aftermath of violent injury & Trauma and injury & Trauma and physical aftermath &  \cellcolor{black}  & Prior experiences with violence and injury \\
        \hline
        Altercation leading to hospital visit & Physical altercation & Physical altercation &   \cellcolor{black} &  \cellcolor{black} \\
        \hline
        Behavior changes after incident & Emotional response to the incident & Pre-incident vs. Post-incident behavior &  \cellcolor{black} &  \cellcolor{black} \\
        \hline
        Feelings on retaliation & Desire for retaliation & Struggle with retaliation & \cellcolor{black}  & Immediate desire for retaliation \\
        \hline
        Influences on the youth & Youth & Youth influence &  \cellcolor{black} & Youth are seen as role models for their peers \\
        \hline
        Masculinity & Masculinity & Masculinity & Masculinity & Defining masculinity \\
        \hline
        Perceived community violence & Neighborhood violence & Community violence & Gun violence & Racialized violence and victimhood \\
        \hline
        Previous incarceration history & Ex-offenders & Past incarceration experience &  \cellcolor{black}  & Multiple stints in prison \\
        \hline
        Reasoning for beef starting & Beef (conflict, dispute, or conflict) & Beef &  \cellcolor{black} &  \cellcolor{black} \\
        \hline
        Sexual activity & Sexual activity & Leisure activities &  \cellcolor{black} &   \cellcolor{black}\\
        \hline
        Substance use & Substance use & Substance use & Substance abuse & Substance use as a way to escape or avoid problems \\
        \hline
    \end{tabular}
        \caption{Initial machine codes that are most semantically similar to the human codes. Darkened cells indicate there was no machine match}
        \label{tab:initialcodes}
    \end{table*}

    \begin{sidewaystable*}[h]
    \centering
    \begin{tabular}{| p{2.5cm} | p{2.5cm}  | p{2.5cm}| p{2.5cm} |p{2.5cm} | p{2.5cm} |p{2.5cm}|p{2.5cm} |p{2.5cm}|}
        \hline
       HC & Best L1B & & Best L8B & & Worst L1B &  & Worst L8B & \\
        & Cluster Match & Semantic Match &  Cluster Match & Semantic Match & Cluster Match & Semantic Match &  Cluster Match & Semantic Match \\
        \hline
        \hline
        Aftermath of violent injury &  Family dynamics & Trauma & Societal norms and expectations & Physical trauma and injury & Get & \cellcolor{black} &  Emotional regulation and self-control & Psychological aftermath of trauma\\
        \hline
        Altercation leading to hospital visit & \cellcolor{black} & \cellcolor{black} & Societal norms and expectations & \cellcolor{black} & Get & \cellcolor{black} & Regret over missed opportunities in life & \cellcolor{black}\\
        \hline
        Behavior changes after incident & \cellcolor{black}  & \cellcolor{black} & Societal norms and expectations &\cellcolor{black} &  \cellcolor{black} &\cellcolor{black} & Peer influence on substance abuse & \cellcolor{black} \\
        \hline
        Feelings on retaliation & \cellcolor{black} & \cellcolor{black} & Physical trauma and injury & \cellcolor{black} & \cellcolor{black} & \cellcolor{black}& \cellcolor{black} & \cellcolor{black}\\
        \hline
        Influences on the youth &\cellcolor{black} &\cellcolor{black} & Role models in masculinity & \cellcolor{black} & Relationship  &\cellcolor{black} & Resilience in the face of adversity & Negative influences on young people\\
        \hline
        Masculinity & Dealing with masculinity and identity & Dealing with masculinity and identity & Racial profiling and police bias & Role models in masculinity & Relationship & \cellcolor{black} & Racism in a post-racial society & Defining masculinity \\
        \hline
        Perceived community violence &\cellcolor{black} & Gun violence & Career development and stability & Community-specific approaches to addressing gun violence & \cellcolor{black} & \cellcolor{black} & \cellcolor{black} & Normalization of violence in communities \\
        \hline
        Previous incarceration history & Empathy & \cellcolor{black} & Relationship dynamics  &\cellcolor{black} & Get & \cellcolor{black}& Family strains and communication breakdowns & \cellcolor{black} \\
        \hline
        Reasoning for beef starting & \cellcolor{black} & \cellcolor{black} & Impact of gun violence on survivors & \cellcolor{black} & \cellcolor{black} &  \cellcolor{black} & \cellcolor{black} & \cellcolor{black} \\
        \hline
        Sexual activity &  \cellcolor{black}& \cellcolor{black} & \cellcolor{black} & \cellcolor{black} & Like & \cellcolor{black} & \cellcolor{black}  \\
        \hline
        Substance use & Lack &  Substance use & Uncertainty and Ambiguity in life experiences & Substance use and abuse & \cellcolor{black} & \cellcolor{black} & Gun violence and trauma in communities &  Struggling with substance addiction \\
        \hline
    \end{tabular}
   \caption{The formal machine codes that the human codes align best with. Darkened cells indicate there was no match.}
    \label{tab:formalcodes}    
    \end{sidewaystable*}

    \textbf{Initial human annotated codes:}
        \begin{itemize}
            \item Use of deescalation tactics 
            \item Feelings of spirituality from incident  
            \item ACEs
            \item  Influence of neighborhood factors on self - peer  
            \item Trauma recidivism 
            \item Influence of neighborhood factors on self - older adults  
            \item Thoughts on racism 
            \item Demographic information 
            \item Thoughts of incident being life changing moment  
            \item  Daily routine 
            \item Symptoms of traumatic stress 
            \item  Current PTSD 
            \item Substance use 
            \item Perpetuation of violence through the system 
            \item Social media use 
            \item  Religion 
            \item Sexual activity 
            \item  Sexuality 
            \item Rooted prejudice 
            \item  Street Codes 
            \item Reasoning for problems escalating or "beefs" starting 
            \item Support System 
            \item Reason for hospital visit/ Diagnosis at hospital 
            \item  Videos of incident 
            \item Reaction to the injury  
            \item  Probation or parole 
            \item Putting in work definition 
            \item  Previous incarceration history 
            \item Perceived community violence 
            \item  Gang involvement (not neighborhood) 
            \item Masculinity  
            \item  Misconception of medical care 
            \item Influences on the youth  
            \item  Altercations with police 
            \item Health care coverage 
            \item  Altercation leading up to hospital visit 
            \item Frequency of healthcare  
            \item  Carrying illegally for protection 
            \item Feelings on retaliation 
            \item Lack of praise for those doing good 
            \item Struggles for black men in society 
        \end{itemize}
      
\textbf{Best Llama-3.2-1B Instruct formal codes (2 topics refused output, 1 null):}                
    \begin{itemize}
            \item Neighborhood 
            \item Physical Discomfort 
            \item Relationships 
            \item Perpetrators 
            \item Substance Use 
            \item Fear 
            \item Lack 
            \item Affirmation 
            \item Good 
            \item Sense of Disconnection 
            \item Post-racial Society 
            \item Perpetrator-Victim Dynamics 
            \item Systemic Inequality 
            \item Stability 
            \item Violence (3) 
            \item Older 
            \item Work 
            \item Identity 
            \item Protection 
            \item Confusion 
            \item Safety 
            \item Care 
            \item Self-preservation 
            \item Police 
            \item Self-defense (2) 
            \item Gang Affiliation Codes: Blood Gangs 
            \item Security 
            \item Anger 
            \item Self-safety 
            \item Gun Violence 
            \item Self-desire 
            \item Internet 
            \item Avoidance 
            \item  Defensiveness 
            \item Situations 
            \item Transporters 
            \item Boys 
            \item Last 
            \item Dealing with Masculinity and Identity 
            \item Continuation 
            \item De-escalation 
            \item  Social Relationships 
            \item Trauma 
            \item  Moment 
            \item Family Dynamics 
            \item  Perceived Threat of Death 
            \item Checkup 
            \item  Nonverbal Communication 
            \item Perceived Societal Expectations 
            \item  Trust in Authority (2) 
            \item Change 
            \item  Education 
            \item Frustration and Anger
            \item  Empathy 
            \item Resilience and Coping Mechanisms 
            \item  Spiritual 
            \item Self-improvement Past 
        \end{itemize}

    \textbf{Best Llama 3.1-8B Instruct formal codes:}
    \begin{itemize}
        \item Societal Norms and Expectations
        \item Relationship with Authority
        \item Violence and Aggression
        \item Career Development and Stability
        \item Resilience and Survival
        \item Danger and Risk in Vulnerable Contexts
        \item Avoidance Tactics in Conversation
        \item Childhood and Coming of Age
        \item Role Models in Masculinity
        \item Racial Profiling and Police Bias
        \item Family Dynamics and Structure
        \item Healthcare Access Barriers
        \item Fear and Anxiety
        \item Emotional Vulnerability
        \item Emotional Expression and Regulation
        \item Gun Violence
        \item Geographic Location and Sense of Place
        \item Time Perception and Transition
        \item Lack of Guidance and Structure for Young Black Men
        \item Public Perception of the Criminal Justice System
        \item Relationship Dynamics
        \item Gun Violence and Victimization
        \item Social Support Systems
        \item Self-Blame and Accountability
        \item Substance Use and Abuse
        \item Uncertainty and Ambiguity in Life Experiences
        \item Validation
        \item Faith and Spirituality
        \item Stigma and Shame in Vulnerable Populations
        \item Autonomy and Self-Determination in Masculinity
        \item Mental Health and Wellbeing of Black Men
        \item Contextualizing Gun Violence
        \item Social Commentary and Moral Judgment
        \item Prioritization of Needs Over Health
        \item Avoidance of Disclosure
        \item Street Culture and Identity
        \item Conflict Resolution Strategies
        \item Impact of Gun Violence on Survivors
        \item Self-Awareness and Personal Growth
    \end{itemize}

\textbf{Worst Llama 3.2-1B Instruct formal codes (3 refused and 2 null):}
        
\begin{itemize}
    \item Health: High Blood Pressure 
    \item  Relationship 
    \item Get 
    \item Like 
    \item Gunshot 
    \item Privilege 
    \item Racial Identity 
\end{itemize}

\textbf{Worst Llama 3.1-8B Instruct formal codes:}
     
\begin{itemize}
    \item Job Insecurity and Limited Opportunities  
    \item Trauma and Mental Health Consequences  
    \item Struggling Young Adults' Motivations and Aspirations  
    \item Defining Masculinity  
    \item Racism in a Post-Racial Society  
    \item Personal Agency and Control  
    \item Psychological Aftermath of Trauma  
    \item Lack of Consequences for Violence in Youth  
    \item Peer Influence on Substance Abuse  
    \item Youth's Relationship with Authority Figures  
    \item Personal Growth and Self-Improvement  
    \item Systemic Inequality and Social Unrest  
    \item Gun Culture and Ownership Practices  
    \item Rediscovery of Purpose and Meaning  
    \item Sense of Belonging to a Close-Knit Community  
    \item Survival and Self-Preservation  
    \item Physical Assault  
    \item Substance Abuse Among Young People  
    \item The Influence of Peer Relationships on Youth  
    \item Resilience in the Face of Adversity  
    \item Physical and Emotional Recovery After Trauma  
    \item Emotional Regulation and Self-Control  
    \item Positive Role Models in the Home and Community  
    \item Impact of Past Experiences on Behavior and Life Choices  
    \item Prioritizing Family Stability  
    \item Social Isolation and Limited Resources  
    \item Masculinity Socialization  
    \item Normalization of Violence in Communities  
    \item Youth's Attitudes Towards Violence  
    \item Delayed Understanding and Acceptance  
    \item Personal Growth and Self-Improvement  
    \item Youth Vulnerability to Gang Involvement  
    \item Reentry and Reincarceration Challenges  
    \item Family Strains and Communication Breakdowns  
    \item Importance of Guidance and Mentorship  
    \item Limited Familiarity with Internet Technology  
    \item Increased Vigilance and Caution in Daily Life  
    \item Uncertainty and Ambivalence  
    \item Conflict De-escalation  
    \item Regret Over Missed Opportunities in Life  
    \item The Search for Identity and Belonging  
    \item Prioritizing Neighborhood Safety  
    \item Loss of Autonomy and Control  
    \item Substance Abuse as a Coping Mechanism  
    \item Gun Violence and Trauma in Communities  
    \item Faith and Spiritual Growth  
    \item Macho Identity  
    \item Side Effects of Opioids  
    \item Relationships and Support Systems  
    \item Negative Influences on Young People  
    \item Struggling with Substance Addiction  
    \item Dependence on Others for Support  
    \item Limited Mobility and Stability  
    \item Limited Job Opportunities  
\end{itemize}

\end{document}